\documentclass[10pt,twocolumn,letterpaper]{article}

\usepackage{cvpr}
\usepackage{times}
\usepackage{epsfig}
\usepackage{graphicx}
\usepackage{amsmath}
\usepackage{amssymb}
\usepackage{array}

\usepackage{multirow}
\usepackage{fancyhdr}
\usepackage{hyperref}
\usepackage{booktabs}
\usepackage{caption}

\usepackage{float}

\newcommand{\PreserveBackslash}[1]{\let\temp=\\#1\let\\=\temp}
\newcolumntype{C}[1]{>{\PreserveBackslash\centering}p{#1}}
\newcolumntype{R}[1]{>{\PreserveBackslash\raggedleft}p{#1}}
\newcolumntype{L}[1]{>{\PreserveBackslash\raggedright}p{#1}}

\cvprfinalcopy % *** Uncomment this line for the final submission

\def\cvprPaperID{****} % *** Enter the CVPR Paper ID here

\begin{document}

\title{Technical Report for ActivityNet Challenge 2022 - Temporal Action Localization}  % **** Enter the paper title here

\author{Shimin Chen\textsuperscript{1}\thanks{These authors contributed equally to this work}
\quad
Wei Li\textsuperscript{1}$^\ast$
\quad 
Jianyang Gu\textsuperscript{1,2}$^\ast$
\quad
Chen Chen\textsuperscript{1}$^\ast$
\quad
Yandong Guo\textsuperscript{1} 
\\
\textsuperscript{1}OPPO Research Institute.
\quad
\textsuperscript{2}Zhejiang University.
\\
{\tt\small \{chenshimin1, liwei19, chenchen, guoyandong\}@oppo.com}
\\
{\tt\small gu\_jianyang@zju.edu.cn}
}

 \twocolumn[{%
 \renewcommand\twocolumn[1][]{#1}%
 \maketitle
 \begin{center}
     \centering
     \includegraphics[height=6cm,width=15cm]{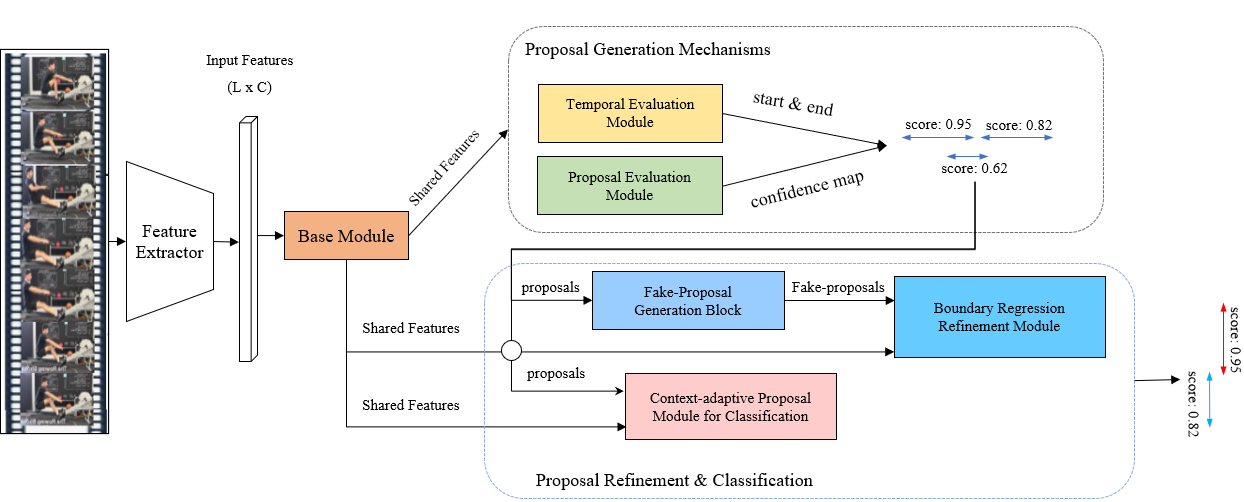}
     \captionof{figure}{Overview of our method. Given an untrimmed video, Faster-TAD can generate proposals and simultaneously (1) refine the boundary and (2) classify the proposal in a context-adaptive way. We construct our Faster-TAD with feature sequences extracted from raw video as inputs.}
     \label{figure:faster-tad framework}
\end{center}%
 }]

\maketitle
\thispagestyle{empty}

%%%%%%%%% BODY TEXT - ENTER YOUR RESPONSE BELOW
\section{Method}
In the task of temporal action localization of ActivityNet-1.3~\cite{caba2015activitynet} datasets, we propose to locate the temporal boundaries of each action and predict action class in untrimmed videos. We first apply VideoSwinTransformer~\cite{liu2021video} as feature extractor to extract different features. Then we apply a unified network following Faster-TAD\cite{chen2022faster} to simultaneously obtain proposals and semantic labels. Last, we ensemble the results of different temporal action detection models which complement each other. Faster-TAD simplifies the pipeline of TAD and gets remarkable performance, obtaining comparable results as those of multi-step approaches.

\subsection{Feature Engineering} \label{sec:fea}

\begin{table}[t]
    \centering
    \caption{Action classification results on validation set of ActivityNet-1.3, measured by Accuracy(\%). $HACS \ Clips400$ set the background of each class as a supplementary class, bring the number of categories in HACS Clips from 200 to 400. $HACS \ Clips200$ stands for commonly used HACS Clips.}
    \label{tab:class_result}
    \scalebox{1.05}{
    \begin{tabular}{c|c|c|c}
    \hline
        Model & Train Data & Top1 ACC& Top5 ACC\\\hline
        CSN &  HACS Clips200 & 89.86 &98.31\\
        CSN & HACS Clips400 & 90.97 &98.43\\
        Swin & HACS Clips200  & 91.16 &98.43\\
        Swin &  HACS Clips400 &91.99 &98.70  \\\hline
        \multicolumn{2}{l|}{ Ensemble Results}&\textbf{93.31} & \textbf{99.11}\\\hline
        
    \end{tabular}}
\end{table}

In order to extract diverse information from untrimmed videos, we make full use of 3 public datasets to pretrain model: Kinetics-700, HACS Clips and AVA-kinetics datasets. We use the pretrained models on Kinetics-700 and HACS Clips to extract features. On the other hand, we train a Slowfast model with AVA-Kinetics dataset to learn action characteristics, and extract features with window size=64 and stride=32. To better make use of the background information of HACS Clips, we set the background of each class as a supplementary class in training, and that change the

\begin{table*}[t]
    \centering
    \caption{Action detection results on validation set of ActivityNet-1.3, measured by mAP(\%) at different tIoU thresholds and the average mAP(\%). $S-N$ stands for Single-Network. $Ensemble \ classifiers$ stands for ensemble video level classification results. $Ensemble \ C-Map$ stands for ensemble confidence map. $CSN-H$ is extracted by CSN model trained on HACS Clips. $CSN-K$ is extracted by CSN model trained on Kinetics700. $Swin-K$ is extracted by VideoSwinTransformer model trained on Kinetics700. $Swin-H$ is extracted by VideoSwinTransformer model trained on HACS Clips. $Slowfast-A$ is extracted by Slowfast model trained on AVA-Kinetics\cite{li2020ava}. $CEL$ stands for cross entropy loss.}
    \label{tab:tal_result}42.
    \scalebox{1.05}{
    \begin{tabular}{c|c|c|c|c|c|c}
    \hline
        Method & Feature & Class-Loss& Ensemble C-Map& S-N &  Video Level Results & Avg  \\\hline
        \multicolumn{7}{l}{Ensemble Video Level Results}\\\hline 
        Faster-TAD &  CSN-H & CEL &\checkmark&$\times$ &\checkmark   & 40.25\\
        Faster-TAD &  Swin-K & CEL &\checkmark&$\times$ &\checkmark  & 40.26\\
        Faster-TAD &  Swin-H & CEL &\checkmark&$\times$ &\checkmark  & 40.28\\
        Faster-TAD & Swin-H+Slowfast-A &Triplet+CEL &\checkmark &$\times$ & \checkmark& \textbf{40.72}  \\\hline
         \multicolumn{7}{l}{Without Ensemble Video Level Results}\\\hline
        Faster-TAD &  CSN-H & CEL & $\times$ &\checkmark & $\times$   & 38.25\\
        Faster-TAD &  CSN-H & CEL &\checkmark &\checkmark & $\times$   & 40.10\\
        Faster-TAD &  Swin-H & CEL &$\times$&\checkmark & $\times$  & 39.10\\
         Faster-TAD &  Swin-H & CEL &\checkmark&\checkmark & $\times$  & 40.33\\
        Faster-TAD & CSN-H+CSN-K & CEL&$\times$ &\checkmark & $\times$  & 39.24 \\
        Faster-TAD & CSN-H+CSN-K & CEL&\checkmark&\checkmark & $\times$  & 40.82 \\
         Faster-TAD & Swin-H+Swin-K &Circle+CEL&$\times$ &\checkmark & $\times$  & 39.59 \\
        Faster-TAD & Swin-H+Slowfast-A & CEL&\checkmark&\checkmark & $\times$ & 40.85\\
        Faster-TAD & Swin-H+Slowfast-A &Triplet+CEL& \checkmark &\checkmark& $\times$ & \textbf{40.92} \\\hline
        \multicolumn{6}{l|}{ Ensemble Results on Val} & \textbf{42.00}\\
        \multicolumn{6}{l|}{ Ensemble Results on Test} & \textbf{44.84}\\\hline
        
    \end{tabular}}
\end{table*}

\noindent  number of categories in HACS Clips from 200 to 400. The results of each atomic classifier are shown in the Table \ref{tab:class_result}.

\subsection{Temporal Action Detection}
\subsubsection{Temporal Proposal Generation}
We apply a Faster-RCNN like network in this temporal action detection task, dubbed Faster-TAD\cite{chen2022faster}. By jointing temporal proposal generation and action classification with multi-task loss and shared features, Faster-TAD simplifies the pipeline of TAD.

As shown in figure \ref{figure:faster-tad framework}, we construct our Faster-TAD with feature sequences extracted from raw video as inputs by VideoSwinTransformer\cite{liu2021video} Extractor. We process the feature sequences with a base module to extract shared features, which consists of a CNN Layer, a Relu Layer, and a GCNeXt\cite{xu2020g} Block. We then exert a Proposal Generation Mechanism to obtain most credible $K$ coarse proposals, where $K$ is 120. Proposals and shared features are further utilized to get more accurate boundaries by Boundary Regression Refinement Module\cite{qing2021temporal}. At the same time, shared features and proposals are employed to get the semantic labels of action instances with Context-Adaptive Proposal Module.

We make some improvements to tackle the challenges in temporal action detection.
Faster-TAD includes Context-Adaptive Proposal Module to adaptively learn the semantic information of proposals by introducing attention mechanism across proposals to whole video and considering context as proximity-category proposals. Then the Fake Proposal based on the ground truth boundary with different offsets improves the Boundary Regression Module. Also, we found feature representation trained on atomic actions is very useful for complex activity detection. We employ Auxiliary-Features Block to adapt to the two streams input, and obtains remarkable performance.

\subsubsection{Proposal Classification}
In order to get clear classification boundaries, we propose to involve metric learning loss functions for explicit constraints of embedded feature distributions. In addition to the commonly utilized cross entropy loss, we adopt 2 metric learning loss functions in total: triplet loss~\cite{schroff2015facenet} and circle loss~\cite{sun2020circle}. In order to explicitly constrain the similarity relationships between positive and negative sample pairs, during the training process, a mini-batch is grouped with $P$ unique categories, each with $K$ samples. As a sample may contain more than 1 category, only the first is taken into consideration at the batch sampling stage.  Metric learning losses aim to form compact clusters for each category. 

For an anchor sample in the mini-batch as $x^i$, whose similarity to positive and negative samples as $s^i_p$ and $s^i_n$, the triplet loss~\cite{schroff2015facenet} can be formulated with:
\begin{equation}
    \mathcal{L}_{tr}=\left[s^i_n-s^i_p+m\right]_+,
\end{equation}
where $m$ represents the margin between clusters, and $[]_+$ stands for $\max(\cdot, 0)$. Triplet loss directly pulls close samples of the same category and pushes away those of different categories. However, as the calculation only involves samples inside the mini-batch, the optimization is easily stuck at local-optima.

Circle loss~\cite{sun2020circle} further introduces weighting factors $\alpha$ and respective margins $\triangle$ for positive and negative sample pairs:
\begin{equation}
    \mathcal{L}_{cr}=\log\left[1+\sum^L_{j=1}\sum^M_{i=1}e^{(\gamma(\alpha^j_n(s^j_n-\triangle_n)-\alpha^i_p(s^i_p-\triangle_p)))}\right].
\end{equation}
In the actual calculation process, the weighting factors are assigned as $\alpha^i_p=[1+m-s^i_p]_+$ and $\alpha^j_n=[s^j_n+m]_+$. The margins are set as $\triangle_p=1-m$ and $\triangle_n=m$. The above loss functions are grouped in multiple ways to produce different TAD models. We employ model ensemble to aggregate the advantages of one another.

\subsection{Ensemble}
In the Chapter \ref{sec:fea} mentioned before, we can generate discriminative features for temporal action detection. In this section, we synthesize the proposal classification results to form the final classification results, and apply soft-NMS \cite{bodla2017soft} to the proposal localization results with different thresholds for different category. Besides, Boundary-Matching confidence map mentioned in BSN\cite{lin2018bsn} enumerates all possible combination of temporal locations, bringing promotion in both efficiency and effectiveness. In this section, we fuse the confidence maps of different models with different weights to further enhance the performance of each single model.
\section{Experiment}
We train our TAD model in a single network, with batch size of 64 on 8 gpus. The learning rate is $6\times{10}^{-4}$ for the first 3 epochs, and is reduced by 10 in epoch 3 and 7. We train the model with total 10 epochs. In inference, we apply Soft-NMS\cite{bodla2017soft} for post-processing, and select the top-M prediction for final evaluation. M is 120.

The results of classification on the val dataset are shown in Table \ref{tab:class_result}. We obtain top1 acc 93.1\% by ensembling video level classification results. The results of TAD on the val dataset are shown in Table~\ref{tab:tal_result}, which measured by TAD metrics mAP(\%) at different tIoU thresholds and the average mAP(\%) as ActivityNet-1.3~\cite{caba2015activitynet}.

{\small
\bibliographystyle{unsrt}
\bibliography{egbib}
}

\end{document}